\documentclass{article}
\usepackage[numbers]{natbib}
\usepackage[T1]{fontenc}
\usepackage[utf8]{inputenc}
\usepackage[final]{arxiv}
\usepackage{amsmath,url}
\usepackage{graphicx}
\usepackage{color}
\usepackage{booktabs}
\usepackage{CJK}
\usepackage{enumitem}
\usepackage{adjustbox} 
\usepackage{pifont}
\usepackage{multicol}
\usepackage{multirow}
\usepackage{amssymb}
\usepackage[table]{xcolor}
\usepackage{tabularx}
\usepackage{subcaption}
\usepackage{xspace}

\newcommand{\ie}{\emph{i.e.}\xspace}
\newcommand{\eg}{\emph{e.g.}\xspace}

\makeatletter
\def\blfootnote{\xdef\@thefnmark{}\@footnotetext}
\makeatother

\title{
{\raisebox{-4pt}{
\includegraphics[width=0.14\linewidth]{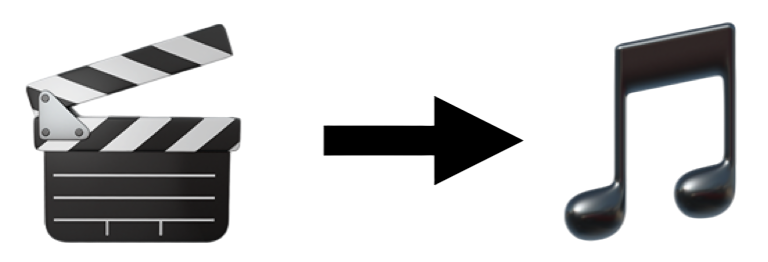} 
}}Vision-to-Music Generation: A Survey}

\author{
\textbf{Zhaokai Wang}\textsuperscript{1}, \
\textbf{Chenxi Bao}\textsuperscript{2}, \
\textbf{Le Zhuo}\textsuperscript{3}, \
\textbf{Jingrui Han}\textsuperscript{4}, \\
\textbf{Yang Yue}\textsuperscript{5}\textbf{,} \
\textbf{Yihong Tang}\textsuperscript{6}\textbf{,} \
\textbf{Victor Shea-Jay Huang}\textsuperscript{2}\textbf{,} \
\textbf{Yue Liao}\textsuperscript{7\dag}\\
[2mm]
\textsuperscript{1}Shanghai Jiao Tong University \quad
\textsuperscript{2}Music Tech Lab, DynamiX \\
\textsuperscript{3}Shanghai AI Laboratory \quad
\textsuperscript{4}Beijing Film Academy \quad
\textsuperscript{5}Tsinghua University \\
\textsuperscript{6}McGill University \quad
\textsuperscript{7}The Chinese University of Hong Kong
\\ 
[2mm]
~\small{\texttt{wangzhaokai@sjtu.edu.cn} \quad \texttt{\{cloudingcxb17,zhuole1025,liaoyue.ai\}@gmail.com}} \\
}

\definecolor{lightblue}{RGB}{54, 116, 230}

\usepackage[colorlinks, linkcolor=red, citecolor=lightblue]{hyperref}

\begin{document}

\maketitle

\blfootnote{\noindent\textsuperscript{\dag}Corresponding author.}

\begin{abstract}

Vision-to-music Generation, including video-to-music and image-to-music tasks, is a significant branch of multimodal artificial intelligence demonstrating vast application prospects in fields such as film scoring, short video creation, and dance music synthesis. 
However, compared to the rapid development of modalities like text and images, research in vision-to-music is still in its preliminary stage due to its complex internal structure and the difficulty of modeling dynamic relationships with video. 
Existing surveys focus on general music generation without comprehensive discussion on vision-to-music.
In this paper, we systematically review the research progress in the field of vision-to-music generation.
We first analyze the technical characteristics and core challenges for three input types: general videos, human movement videos, and images, as well as two output types of symbolic music and audio music.
We then summarize the existing methodologies on vision-to-music generation from the architecture perspective.
A detailed review of common datasets and evaluation metrics is provided.
Finally, we discuss current challenges and promising directions for future research. 
We hope our survey can inspire further innovation in vision-to-music generation and the broader field of multimodal generation in academic research and industrial applications.
To follow latest works and foster further innovation in this field, we are continuously maintaining a GitHub repository at {\small \url{https://github.com/wzk1015/Awesome-Vision-to-Music-Generation}}.

\end{abstract}

\begin{figure*}[t]
  \centering
  \includegraphics[width=\linewidth]{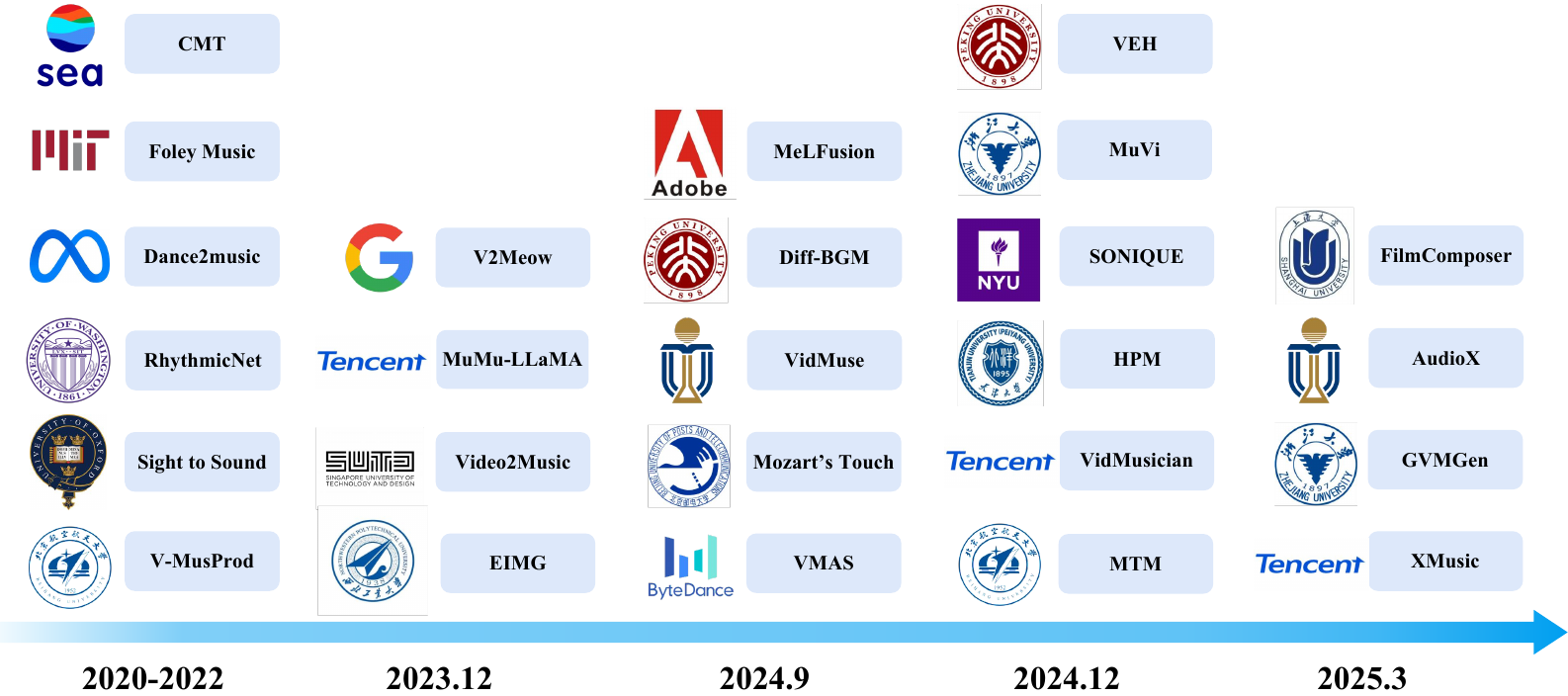}
  \vspace{1mm}
   \caption{Timeline of representative works in vision-to-music generation.}
   \label{fig:timeline}
\end{figure*}

\section{Introduction}\label{sec:introduction}

Recent advances in multimodal artificial intelligence have witnessed substantial progress in generating and understanding content for modalities like text, images, video, and speech~\cite{GPT-4, video_llama, mono_internvl, emu3,  phyworld}. 
Music generation, as an important part of this multimodal ecosystem, has also seen remarkable development. Among the various music generation tasks (\eg unconditional music generation~\cite{audiolm, compound_word_transformer} and text-to-music generation~\cite{musiclm, MusicGen}), \textbf{vision-to-music}, including \textbf{video-to-music} and \textbf{image-to-music generation}, has garnered particular interest due to its practical applications in film scoring, short video platforms, and music accompaniment. For general users, automatically generated background music can alleviate copyright concerns and reduce the time spent searching for suitable music. For professional composers, AI-assisted music composition can streamline the iterative process of matching a score to visual content, expediting the communication cycle with directors and producers.

Despite this demand, the development of vision-to-music remains relatively preliminary. For academic research, the inherent challenges ranging from aligning rich visual cues with musical structure to handling the multifaceted nature of music generation contribute to the task's high complexity, making it more difficult than the common text-to-music task~\cite{MusicGen, musiclm, StableAudioOpen}.
For instance, Videos and music may not have a one-to-one relationship. In fact, a single video can even correspond to various emotions and different styles of music, which depends on the creator's intentions and the subjective feelings of the audience. This correspondence is more complex compared to traditional multimodal generation tasks such as image captioning and text-to-image generation.
Although a growing number of works have emerged in recent years~\cite{CMT, V2Meow, V-MusProd, MuMu-LLaMA, VidMuse, filmcomposer}, they are still far from meeting the diverse requirements of real-world scenarios. 

For industrial integration, while other AI-generated content (AIGC) fields, such as text-to-image~\cite{esser2024scaling,zhuolumina,chenpixart} and text-to-video~\cite{movie_gen,kong2024hunyuanvideo}, have experienced rapid adoption in both professional and consumer contexts, vision-to-music systems have yet to see broad industrial deployment, with only pilot products like Tianpuyue AI\footnote{https://www.tianpuyue.cn/video2music}. The unique demands of film scoring, which often require precise emotional and temporal synchronization with visual storytelling, heighten the difficulty of achieving robust and artistically consistent results through AI methods.

\begin{figure*}[t]
    \centering
    \includegraphics[width=\linewidth]{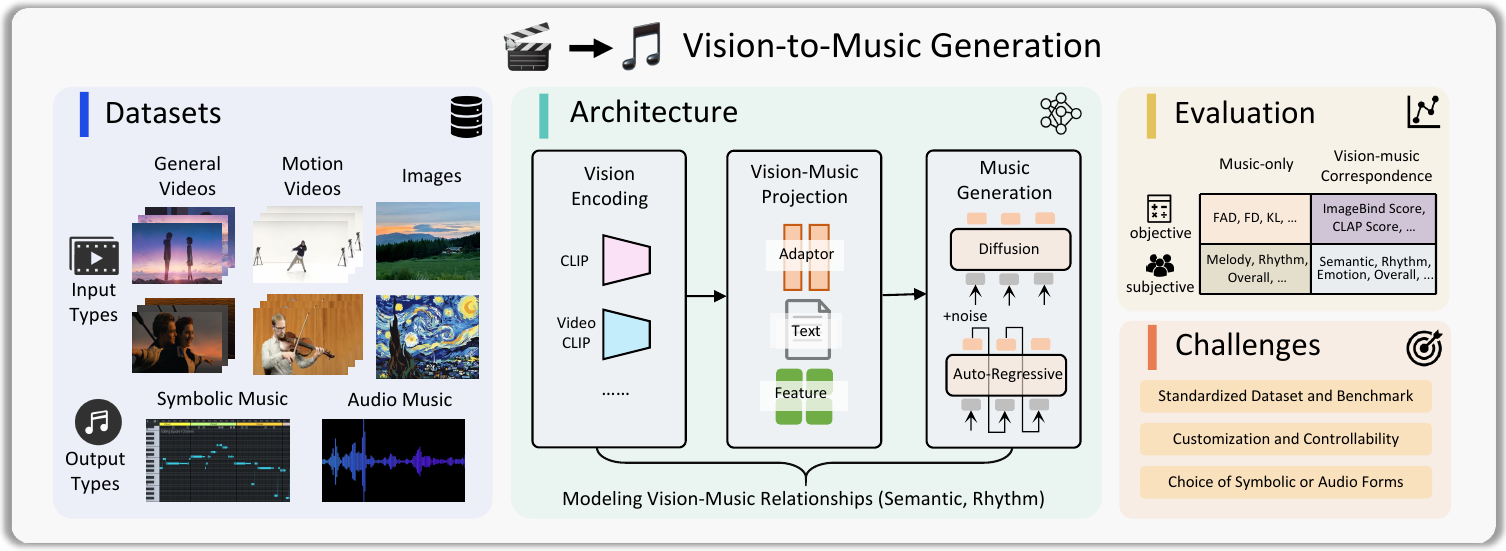}
    {
    \caption{Overview of vision-to-music generation.}
    
    \label{fig:overview}
    }
\end{figure*}

Although some existing works provide reviews on general music generation~\cite{survey_music_foundation, survey_symbolic_music, survey_deep_music, survey_music_composition, survey_ai_music}, but there lack surveys focusing on the vision-to-music generation task. Given the above gaps, we aim to provide a comprehensive survey on vision-to-music generation.
We provide a timeline of representative works in Fig.~\ref{fig:timeline}, and an overview of vision-to-music generation in Fig.~\ref{fig:overview}. 

The subsequent sections of this paper are organized as follows: Sec.~\ref{sec:fundamentals} introduces the fundamentals of vision-to-music generation, analyzing the technical characteristics and core challenges of three major scenarios: general videos, human movement videos, and images. Sec.~\ref{sec:methods} reviews current vision-to-music methods, comparing innovations and limitations in vision encoding, vision-music projection, and music generation module design. Sec.~\ref{sec:datasets} discusses recent vision-to-music datasets. Sec.~\ref{sec:evaluation} introduces evaluation metrics, categorized their purposes (music-only and vision-music correspondence) and approaches (objective and subjective).
Sec.~\ref{sec:challenges} discusses the current research status and existing challenges. 

Through this work, we aspire to inspire further innovation in vision-to-music generation and the broader field of AI music generation and multimodal learning communities, driving progress in both academic research and industrial applications of vision-to-music generation. In addition to the survey, we have established a GitHub repository with latest papers in this field at \url{https://github.com/wzk1015/Awesome-Vision-to-Music-Generation}. We will actively maintain it and incorporate new research as it emerges.

\section{Fundamentals}
\label{sec:fundamentals}

In the broad multimodal research community~\cite{GPT-4, InternVL1.5, sparkle, yue2024deer, piip_v2, tide, kong2024hunyuanvideo}, music is often treated as a subset of audio~\cite{audiolm,AudioLDM,movie_gen,vatt,LORIS, codi, NeXTGPT}. However, unlike general audio which may include background noise, speech, or sound effects, music embodies intricate internal structures and richness of information, including harmony, counterpoint, and instrumentation. These complexities make it essential to consider music as an independent modality, which sets the stage for exploring vision-to-music generation. 

When delving into this specific area, we first need to recognize the unique relationship between visual input and musical output.
We analyze the characteristics of three input types in vision-to-music generation: \emph{general videos}, \emph{human movement videos}, and \emph{images}, and two output types: \emph{symbolic music} and \emph{audio music}. This categorization helps us better understand the current state and challenges of the field.

\subsection{Input Types}

Vision-to-music can be divided into three input types, which affect the nature of the approaches employed.

\noindent\textbf{General Videos.}
   This category includes methods that take general video inputs and generate symbolic or audio music outputs. 
   This includes a wide range of video contents, such as natural landscapes, films, sports, animations, etc. Techniques in this category typically focus on extracting features like motion, color, or visual semantics to create music that aligns with the visual narrative.
   However, most general video-to-music approaches do not clearly distinguish between various application scenarios, and they overlook the differences in compositional logic across different contexts. For instance, soundtracks for short videos require a very stable structure, whereas film scoring tends to have a looser and more free-form structure. Low quality video-to-music data is abundant and easy to collect, but the sources are very diverse with uneven quality, lacking effective categorization and alignment.

\noindent\textbf{Images.}
   Approaches in this domain focus on transforming static images into music. Since images lack temporal semantics and rhythm, these methods usually only need to focus on the overall style, and there is no strict requirement for the duration of the generated music. The application scenarios for image-to-music are not as extensive as video-to-music, but they include functionalities like creating musical memories for photo albums. This type of pair data is easy to collect, but its inherent correlation may not be very strong.

\noindent\textbf{Human Movement Videos.}
   These videos typically include dance, sports, instrument performances, and other human movements. For instrument performance videos, where humans play music instruments but the audio is removed, the music is determined by the input video to some extent, and the generation process is similar to reconstructing music from the silent videos. For dance, sports, and other human movements, they emphasize rhythmic alignment (especially local rhythm) more than general videos, while semantic constraints are generally weaker, requiring only overall style matching. Therefore, extracted 2D/3D keypoints representing human motion are often directly used as inputs instead of raw videos.

\clearpage

The differences among these tasks have not been fully explored by the current machine learning community. For instance, most works do not differentiate between film scoring and other types of music composing (both falling under general video-to-music), even though they have essential distinctions. From the perspective of the music community, film scoring has the following characteristics:  Film scoring goes beyond merely matching a visual's style or mood. It is deeply intertwined with narrative structure and character development. Unlike tasks such as recovering music from instrumental performance videos or generating music for dance videos, which primarily focus on rhythm, synchronization, or aesthetic alignment, film scoring demands that the music not only mirror the emotional and visual cues of a scene but also advance the storytelling. This involves a highly collaborative, iterative process with directors and editors to ensure that musical motifs (\emph{leitmotifs}), dynamic transitions, and thematic progressions accurately reflect the unfolding narrative and subtext.

Besides, a major difference among the three input types lies in the rhythmic relationships: the rhythm of general videos is jointly determined by the semantic content of the images (such as whether the scene is intense and exciting), the amplitude of movements, scene transitions, etc. For human movement, the music rhythm generally has a direct correspondence with human body movements. As for images, they usually do not include a rhythm relationship or only provide a very implicit global rhythm based on the semantic content of the images. In addition, the length of a video determines the length of the corresponding music, while an image places no restrictions on the length of the music.

In the remaining sections, we will mainly focus on general videos and images, while paying relatively less attention to human movement videos. This is because their semantic association with music is not strong, and they often directly use 2D body keypoints as video features, which is discusses in Sec.~\ref{sec:methods}. Their application scenarios are also relatively limited.

\subsection{Output Types}
From the perspective of music modality, there are two primary types of music to consider:  

\noindent\textbf{Symbolic Music.}
   Symbolic music is represented as discrete elements like notes, chords, or sequences of musical symbols~\cite{survey_symbolic_music}. Most early vision-to-music methods are symbolic~\cite{CMT, V-MusProd, Video2music, Diff-BGM}. Symbolic music can incorporate music theory, such as chords, and generate longer pieces with good controllability. However, the limited data availability restricts its scalability to large models, and the expressive and emotional depth is constrained by soundfonts. 
   This is similar to why end-to-end speech synthesis is prefered over generating text and then converting it to audio.

\noindent\textbf{Audio Music.}
   Such methods aim to generate music in its audio form~\cite{V2Meow, MuMu-LLaMA, VidMuse, MuVi, VidMusician, MTM, GVMGen, filmcomposer}, often employing generative models such as transformers~\cite{Transformer}, VAEs, GANs~\cite{GAN}, or diffusion models~\cite{DDPM} to synthesize realistic sound from the visual input. Audio music benefits from large-scale datasets for training, enabling end-to-end generation with rich expressiveness and performance. However, audio music lacks controllability, and the generated music is typically shorter (usually under 20 seconds) due to sampling rate limitations.

We provide further discussion on the trade-off between symbolic and audio music in Sec.~\ref{sec:challenges}.

\section{Methods}
\label{sec:methods}

In this section, we discuss the existing works on vision-to-music generation. 
We begin by categorizing the methods based on the task classifications as outlined in Sec.~\ref{sec:fundamentals}. Subsequently, we analyze these methods from perspectives such as architecture and vision-music relationships. 
We summarize vision-to-music generation methods in Tab.~\ref{tab:methods}.

\newpage

\subsection{Tasks}
We begin by categorizing the methods based on the input types outlined in Sec.~\ref{sec:fundamentals}.

\noindent\textbf{General Video-to-Music.}
Early video-to-music works typically focused on generating music from instrumental performances~\cite{Audeo, multi_instrument_net, FoleyMusic, sight_to_sound} or dance videos~\cite{rhythmicnet, D2M-GAN, CDCD}. For general videos, the earliest works were primarily based on recommendation/retrieval~\cite{hong2018cbvmr, li2019query} rather than generation. 
Early general video-to-music methods were usually symbolic. 
CMT~\cite{CMT} is a pioneering work in general video-to-music. Due to the lack of paired data for training, they proposed a paradigm where pure music data is used during training, and video features are mapped to music features during inference, with the correspondence primarily based on rhythm. V-MusProd~\cite{V-MusProd} and Video2music~\cite{Video2music} collected video-music paired symbolic datasets, thereby incorporating semantic relationships. Diff-BGM~\cite{Diff-BGM} was the first to introduce diffusion models for video-to-symbolic music generation. V2Meow~\cite{V2Meow} also uses both rhythmic and semantic correspondences and was the first to generate background music in audio form. With the development of multimodal large models and audio-form music generation~\cite{StableAudioOpen, MusicGen}, a large number of general video-to-music works have emerged in the past few years~\cite{MuMu-LLaMA, MozartTouch, VidMuse, VMAS, MuVi, HPM, VidMusician, MTM, GVMGen}, most of which focus on audio modality.

\noindent\textbf{Image-to-Music.}
Early image-to-music works were also primarily symbolic~\cite{Synesthesia, MuSyFI, Vis2Mus, BGT}, where models analyze color, texture, and semantic content to generate music. 
Notably, these methods usually produce symbolic music outputs, where the connection between the image and music is more abstract compared to video-based methods. 
The methodologies often leverage pretrained vision models to extract key image features, which are then mapped to musical elements. 
More recently, MuMu-LLaMA~\cite{MuMu-LLaMA, M2UGen} introduced large language models for image-to-music, video-to-music, and text-to-music generation, representing early audio-form image-to-music works. Methods like MeLFusion~\cite{MeLFusion}, M2M-Gen~\cite{M2M-Gen}, and MTM~\cite{MTM} propose generating audio music from multiple modalities (video, image, and text).

\noindent\textbf{Human Movement Video-to-music.}
For dance or sports videos, existing methods focus on extracting rhythmic patterns from dance videos and mapping them to musical rhythm generation~\cite{rhythmicnet, D2M-GAN, CDCD, LORIS, liu2024let, dance2music, dance2music_ti, dancecomposer}. For music performance videos, current methods learn to reconstruct the original music from the silent videos~\cite{multi_instrument_net, Audeo, FoleyMusic, sight_to_sound}.

\begin{table*}[t]
\renewcommand\arraystretch{1.2}
\centering
\caption{Methods for vision-to-music generation. Sem: Semantics. Rhy: Rhythm. AR: Auto-regressive. Diff.: Diffusion. 
}
\vspace{-1mm}
\label{tab:methods}
\resizebox{\textwidth}{!}{
\begin{tabular}{lllllllll}
    \toprule
    \multirow{2}{*}{Method} & \multirow{2}{*}{Date} & \multirow{2}{*}{Input Type} & \multirow{2}{*}{Modality} & Music & Vision-Music & \multirow{2}{*}{Vision Encoding} & Vision-Music & \multirow{2}{*}{Music Generation} \\
    & & & & Length & Relationships & & Projection & \\
    
    \midrule
    \noalign{\vskip -1mm}
    \rowcolor{gray!10} \multicolumn{9}{l}{$\blacktriangledown$ \emph{General Videos and Images:}}   \\

    CMT~\cite{CMT} & 2021/11 & General Video & Symbolic & 3min & Rhy & - & Elements & AR (CP~\cite{compound_word_transformer}) \\ 
    V-MusProd~\cite{V-MusProd} & 2022/11 & General Video & Symbolic & 6min & Sem, Rhy & CLIP2Video~\cite{Clip2video}, Histogan~\cite{Histogan} & Feature & AR (CP~\cite{compound_word_transformer})\vspace{0.03mm} \\

    \multirow{2}{*}{V2Meow~\cite{V2Meow}} & \multirow{2}{*}{2023/05} & \multirow{2}{*}{General Video} & \multirow{2}{*}{Audio} & \multirow{2}{*}{10sec} & \multirow{2}{*}{Sem, Rhy} & CLIP~\cite{CLIP}, I3D Flow~\cite{I3D}, & \multirow{2}{*}{Feature} & \multirow{2}{*}{AR} \\ 
    & & & & & & ViT-VQGAN~\cite{ViT-VQGAN} & & \\
    
    MuMu-LLaMA~\cite{M2UGen} & \multirow{2}{*}{2023/11} & \multirow{2}{*}{General Video, Image} & \multirow{2}{*}{Audio} & \multirow{2}{*}{30sec} & \multirow{2}{*}{Sem} & \multirow{2}{*}{ViT~\cite{ViT}, ViViT~\cite{ViViT}} & \multirow{2}{*}{Adapter} & \multirow{2}{*}{AR (LLaMA2~\cite{Llama2})} \\ 
     (M$^2$UGen~\cite{MuMu-LLaMA}) & & & & & & & & \vspace{0.03mm} \\
    
    Video2Music~\cite{Video2music} & 2023/11 & General Video & Symbolic & 5min & Sem, Rhy & CLIP~\cite{CLIP} & Feature & AR \\ 
    EIMG~\cite{EIMG} & 2023/12 & Image & Symbolic & 15sec & Sem & ALAE~\cite{ALAE}, $\beta$-VAE~\cite{betaVAE}, VQ-VAE~\cite{VQ-VAE} & Adapter & VAE (FNT~\cite{FNT}, LSR~\cite{LSR}) \\ 
    Diff-BGM~\cite{Diff-BGM} & 2024/05 & General Video & Symbolic & 5min & Sem & VideoCLIP~\cite{VideoCLIP} & Feature & Diff. (Polyffusion~\cite{Polyffusion}) \\ 
    Mozart's Touch~\cite{MozartTouch} & 2024/05 & General Video, Image & Audio & 10sec & Sem & BLIP~\cite{BLIP} & Text & AR (MusicGen~\cite{MusicGen}) \\ 
    MeLFusion~\cite{MeLFusion} & 2024/06 & Image & Audio & 10sec & Sem & DDIM~\cite{DDIM} + T2I LDM~\cite{StableDiffusion} & Feature & Diff. \\ 
    VidMuse~\cite{VidMuse} & 2024/06 & General Video & Audio & 20sec & Sem & CLIP~\cite{CLIP} & Adapter & AR (MusicGen~\cite{MusicGen}) \\ 
    S2L2-V2M~\cite{S2L2-V2M} & 2024/08 & General Video & Audio & 10sec & Sem & Enhanced Video Mamba & Adapter & AR (LLaMA2~\cite{Llama2}) \\ 
    VMAS~\cite{VMAS} & 2024/09 & General Video & Audio & 10sec & Sem, Rhy & Hiera~\cite{Hiera} & Feature & AR \\ 
    MuVi~\cite{MuVi} & 2024/10 & General Video & Audio & 20sec & Sem, Rhy & VideoMAE V2~\cite{videomae_v2} & Adapter & Diff. (DiT~\cite{DiT}) \\ 
    SONIQUE~\cite{SONIQUE} & 2024/10 & General Video & Audio & 20sec & Sem, Rhy & Video-LLaMA~\cite{Video-llama}, CLAP~\cite{CLAP} & Text & Diff. (Stable Audio~\cite{evans2024fast}) \\ 
    VEH~\cite{VEH} & 2024/10 & General Video & Symbolic & 30sec & Sem & VideoChat~\cite{Videochat} & Text  & AR (T5~\cite{T5}) \\ 
    M2M-Gen~\cite{M2M-Gen} & 2024/10 & Image (Manga) & Audio & 1min & Sem & CLIP~\cite{CLIP}, GPT-4~\cite{GPT-4} & Text & AR (MusicLM~\cite{MusicCaps}) \\ 
    HPM~\cite{HPM} & 2024/11 & General Video & Audio & 10sec & Sem & CLIP~\cite{CLIP}, TAVAR~\cite{TAVAR}, WECL~\cite{WECL} & Feature & Diff. (AudioLDM~\cite{AudioLDM}) \\ 
    VidMusician~\cite{VidMusician} & 2024/12 & General Video & Audio & 30sec & Sem, Rhy & CLIP~\cite{CLIP}, T5~\cite{T5} & Adapter & AR (MusicGen~\cite{MusicGen}) \\ 
    MTM~\cite{MTM} & 2024/12 & General Video, Image & Audio & 30sec & Sem & InternVL2~\cite{InternVL1.5} & Text  & Diff. (Stable Audio Open~\cite{StableAudioOpen}) \\ 
    XMusic~\cite{XMusic} & 2025/01 & General Video, Image & Symbolic & 20sec & Sem, Rhy & ResNet~\cite{ResNet}, CLIP~\cite{CLIP} & Elements & AR (CP~\cite{compound_word_transformer}) \\ 
    GVMGen~\cite{GVMGen} & 2025/01 & General Video & Audio & 15sec & Sem & CLIP~\cite{CLIP} & Adapter & AR (MusicGen~\cite{MusicGen}) \\ 
    AudioX~\cite{audiox} & 2025/03 & General Video & Audio & 10sec & Sem & CLIP~\cite{CLIP} & Feature & Diff. (Stable Audio Open~\cite{StableAudioOpen}) \\
    
    \multirow{2}{*}{FilmComposer~\cite{filmcomposer}} & \multirow{2}{*}{2025/03} & \multirow{2}{*}{General Video} & \multirow{2}{*}{Audio} & \multirow{2}{*}{15sec} & \multirow{2}{*}{Sem, Rhy} & Controllable Rhythm Transformer,  & \multirow{2}{*}{Text} & \multirow{2}{*}{AR (MusicGen~\cite{MusicGen})} \\
    & & & & & & GPT-4v~\cite{GPT-4}, Motion Detector & & \\

    \midrule
    \noalign{\vskip -1mm}
    \rowcolor{gray!10} \multicolumn{9}{l}{$\blacktriangledown$ \emph{Human Movement Videos:}}   \\
        
    Audeo~\cite{Audeo} & 2020/06 & Performance Video & Symbolic & 30sec & Rhy & ResNet~\cite{ResNet} & Feature & GAN \\
    Foley Music~\cite{FoleyMusic} & 2020/07  & Performance Video &  Symbolic & 10sec & Rhy & 2D Body Keypoints & Feature & AR \\
    Multi-Instrucment Net~\cite{multi_instrument_net} & 2020/12 & Performance Video & Audio & 10sec & Rhy & 2D Body Keypoints & Feature & VAE \\
    
    RhythmicNet~\cite{rhythmicnet} & 2021/06 & Dance Video & Symbolic & 10sec & Rhy & 2D Body Keypoints & Feature & AR (REMI~\cite{remi}) \\
    Dance2Music~\cite{dance2music} & 2021/07 & Dance Video & Symbolic & 12sec & Rhy & 2D Body Keypoints & Feature & AR \\
    D2M-GAN~\cite{D2M-GAN} & 2022/04 & Dance Video & Audio & 2sec & Rhy & 2D Body Keypoints, I3D~\cite{carreira2017quo} & Feature & GAN \\
    CDCD~\cite{CDCD} & 2022/06 & Dance Video & Audio & 2sec & Rhy & 2D Body Keypoints, I3D~\cite{carreira2017quo} & Feature & Diff. \\
    LORIS~\cite{LORIS} & 2023/05 & Movement Video & Audio & 50sec & Rhy & 2D Body Keypoints, I3D~\cite{carreira2017quo} & Feature & Diff. \\
    VisBeatNet~\cite{liu2024let} & 2024/01 & Dance Video & Symbolic & Realtime & Rhy & 2D Body Keypoints & Feature & AR \\
    UniMuMo~\cite{unimumo} & 2024/10 & Dance Video & Audio & 10sec & Rhy & 2D Body Keypoints & Feature & Diff. \\

\bottomrule
\end{tabular}
}
\end{table*}

\subsection{Architecture}
The architecture of vision-to-music systems can be broken down into three major components: vision encoding, vision-music projection, and music generation.
Each part plays a crucial role in transforming visual data into musical content.

\noindent\textbf{Vision Encoding.} This stage is focused on extracting features from the input video or image. 
The vision encoder typically uses pretrained convolutional neural networks (CNNs)~\cite{ResNet} or transformers~\cite{Transformer} to capture visual semantic information and other relevant features. 
A commonly used vision encoder is CLIP~\cite{CLIP}, which is pretrained on massive image-text pairs to achieve open-domain visual understanding capabilities. Video understanding backbones~\cite{ViViT, Clip2video, VideoCLIP, Videochat, videomae_v2, Video-llama, InternVL1.5} are also used to extract spatiotemporal features. 
Some also use additional encoders for motion information (\eg I3D Flow~\cite{I3D} in V2Meow~\cite{V2Meow}), color information (\eg Histogan~\cite{Histogan} in V-MusProd~\cite{V-MusProd}, TAVAR~\cite{TAVAR} in HPM~\cite{HPM}), emotion information (\eg WECL~\cite{WECL} in HPM~\cite{HPM}), or intermediate text features (\eg T5~\cite{T5} in VidMusician~\cite{VidMusician}). 
For human movement videos, it is important to extract motion features for rhythmic alignment,~\emph{e.g.}, directly calculating first-order difference from human keypoints as motion velocity~\cite{rhythmicnet,LORIS}, or using pre-trained motion encoder~\cite{V2Meow, D2M-GAN}.

\noindent\textbf{Vision-Music Projection.} This component involves mapping the visual features into the music space. Most methods directly use the visual features as the input of the music generation model, or through simple cross-attention mechanisms~\cite{V-MusProd, Diff-BGM, VidMusician, MeLFusion}. Some methods design specialized adapters~\cite{VidMuse, MuVi, MuMu-LLaMA} for better feature alignment, \eg to capture temporal-related or local features. Besides using feature-based mapping, some studies suggest using text as an intermediate representation of the visual features~\cite{MTM, MozartTouch, SONIQUE, filmcomposer} and subsequently utilizing text-to-music models for music generation. Some symbolic music generation methods use symbolic elements as the vision-music mapping~\cite{CMT, XMusic}.

\noindent\textbf{Music Generation.} Once the visual and music features have been aligned, the next step is to generate the musical output. This stage can be tackled using auto-regressive or diffusion-based generative models. 
Auto-regressive models can be used for both symbolic (\eg CMT~\cite{CMT}, V-MusProd~\cite{V-MusProd}, XMusic~\cite{XMusic} are based on compound word transformers~\cite{compound_word_transformer}) and audio music generation (\eg MusicGen~\cite{MusicGen} in \cite{MozartTouch, VidMuse, VMAS, VidMusician, GVMGen}). Diffusion models~\cite{DDPM, DiT} can be used for symbolic (\eg Polyffusion~\cite{Polyffusion} in Diff-BGM~\cite{Diff-BGM}) music to directly generate piano rolls, but mostly for audio music (\eg AudioLDM~\cite{AudioLDM} in HPM~\cite{HPM}, Stable Audio Open~\cite{StableAudioOpen} in MTM~\cite{MTM}).

\subsection{Vision-Music Relationships}
Vision-music relationships establish the correspondence between videos and music. Unlike the vision-music projection discussed in the previous section (which focuses on the architecture), the relationships discussed here focus on the overall correspondence between input and output. These relationships can be broadly classified into two categories: semantic relationships and rhythmic relationships.

\noindent\textbf{Semantic Relationships.} This type of relationship focuses on how visual elements (such as color, objects, or scenes) relate to musical components (such as mood, melody, or chords). 
For music performance videos, the music is determined by the video instead of a general and implicit semantic relationship~\cite{FoleyMusic,Audeo}. For dance and movement videos, the semantics in the video is not utilized.
Symbolic methods like V-MusProd~\cite{V-MusProd}, Video2music~\cite{Video2music}, and XMusic~\cite{XMusic} explicitly define semantic, color, and emotion relationships and use different pretrained models for extraction. This multi-attribute extraction has unique advantages for symbolic music, as it allows explicit definition of mapping rules between video features and musical attributes for control. 
Recent audio-based methods generally use a single vision encoder to extract semantic features. These semantic features are usually global and insensitive to semantic changes within the video.
Some methods~\cite{VidMuse,GVMGen,MuVi} also design special modules to enhance local semantic correspondence.
Methods like VidMuse~\cite{VidMuse}, GVMGen~\cite{GVMGen}, and MuVi~\cite{MuVi} design special modules to enhance local semantic correspondence.
However, for most audio methods generating 10-second music, the concept of ``local'' may not have a significant impact.

\noindent\textbf{Rhythmic Relationships.} Rhythmic relationships mainly refer to the correspondence between the rhythm of the video (\eg local movements, scene transitions, global video rhythm) and the rhythm of the music (\eg local beats, global tempo). For human movement videos, such as dance or instrument playing, rhythmic relationships become significant, especially the correspondence between local rhythm and human movements. 
For general videos, early work CMT~\cite{CMT} proposed a rule-based rhythmic correspondence based on optical flow. While it can generate rhythmically strong music for many videos, for videos with large motion amplitudes, the music generated by CMT lacks overall rhythmic consistency and may sound slightly chaotic.  Considering the high computational cost of optical flow, V-MusProd~\cite{V-MusProd} adopted RGB-Diff and focused on global rhythm (\ie tempo and global timing encoding). For audio music, explicit rhythm control is more challenging. VMAS~\cite{VMAS} extracts video and music beats using optical flow and onset detection and uses an autoregressive loss for explicit rhythm alignment. V2Meow~\cite{V2Meow} uses optical flow to extract per-frame rhythm features to establish implicit rhythmic correspondence. 
Recent works mostly do not consider rhythm information or use frame-by-frame semantic features to implicitly provide local rhythm correspondence~\cite{VidMuse, VidMusician}, which is not prominent in the generated music. 
Similar to semantic relationships, for most 10-second audio methods, "local" is not a significant concept. 
In methods that use text for vision-music projection~\cite{MTM, SONIQUE}, the video content is used to generate requirements for musical rhythm, such as the rhythm of each scene or the overall tempo.

\section{Datasets}
\label{sec:datasets}

\begin{table*}[t]
\renewcommand\arraystretch{1.15}
\centering
\caption{Datasets for vision-to-music generation. }
\vspace{-1mm}
\label{tab:datasets}
\resizebox{\textwidth}{!}{
    \begin{tabular}{lllllllll}
        \toprule
        \multirow{2}{*}{Dataset} & \multirow{2}{*}{Access} & \multirow{2}{*}{Date} & \multirow{2}{*}{Source} & \multirow{2}{*}{Modality} & \multirow{2}{*}{Size} & Total Length & Avg. Length & \multirow{2}{*}{Annotations} \\
        & & & & & & (hour) & (second) & \\
        
        \midrule
        \noalign{\vskip -1mm}
        \rowcolor{gray!10} \multicolumn{9}{l}{$\blacktriangledown$ \emph{General Videos:}}   \\
        HIMV-200K~\cite{HIMV-200K} & \href{https://github.com/csehong/VM-NET/tree/master/data}{Link} & 2017/04 & Music Video (Youtube-8M~\cite{Youtube-8M}) & Audio & 200K & - & - & - \\
        MVED~\cite{MVED} & \href{https://github.com/yagyapandeya/Music_Video_Emotion_Dataset}{Link} & 2020/09 & Music Video & Audio & 1.9K & 16.5 & 30 & Emotion\vspace{0.03mm} \\
        
        \multirow{2}{*}{SymMV~\cite{V-MusProd}} & \multirow{2}{*}{\href{https://github.com/zhuole1025/SymMV/tree/main/dataset}{Link}} & \multirow{2}{*}{2022/11} & \multirow{2}{*}{Music Video} & \multirow{2}{*}{MIDI, Audio} & \multirow{2}{*}{1.1K} & \multirow{2}{*}{76.5} & \multirow{2}{*}{241} & Lyrics, Genre, Chord, \\
        & & & & & & & & Melody, Tonality, Beat\vspace{0.03mm} \\
        
        MV100K~\cite{V2Meow} & - & 2023/05 & Music Video (Youtube-8M~\cite{Youtube-8M}) & Audio & 110K & 5000 & 163 & Genre\vspace{0.03mm} \\
        
        \multirow{2}{*}{MusicCaps~\cite{MusicCaps}} & \multirow{2}{*}{\href{https://www.kaggle.com/datasets/googleai/musiccaps}{Link}} & \multirow{2}{*}{2023/01} & \multirow{2}{*}{Diverse Videos (AudioSet~\cite{AudioSet})} & \multirow{2}{*}{Audio} & \multirow{2}{*}{5.5K} & \multirow{2}{*}{15.3} & \multirow{2}{*}{10} & Genre, Caption, Emotion, \\
        & & & & & & & & Tempo, Instrument, Rhythm, ...\vspace{0.03mm} \\
        
        EmoMV~\cite{EmoMV} & \href{https://github.com/ivyha010/EmoMV}{Link} & 2023/03 & Music Video (MVED~\cite{MVED}, AudioSet~\cite{AudioSet}) & Audio & 6K & 44.3 & 27 & Emotion\vspace{0.03mm} \\
        MUVideo~\cite{MuMu-LLaMA} & \href{https://huggingface.co/datasets/M2UGen/MUVideo}{Link} & 2023/11 & Diverse Videos (Balanced-AudioSet~\cite{AudioSet}) & Audio & 14.5K & 40.3 & 10 & Instructions\vspace{0.03mm} \\
        
        \multirow{2}{*}{MuVi-Sync~\cite{Video2music}} & \multirow{2}{*}{\href{https://zenodo.org/records/10057093}{Link}} & \multirow{2}{*}{2023/11} & \multirow{2}{*}{Music Video} & \multirow{2}{*}{MIDI, Audio} & \multirow{2}{*}{784} & \multirow{2}{*}{-} & \multirow{2}{*}{-} & Scene Offset, Emotion, Motion, Semantic, \\ 
        & & & & & & & &  Chord, Key, Loudness, Density, ...\vspace{0.03mm} \\
        
        BGM909~\cite{Diff-BGM} & \href{https://github.com/sizhelee/Diff-BGM}{Link} & 2024/05 & Music Video & MIDI & 909 & - & - & Caption, Style, Chord, Melody, Beat, Shot \\
        V2M~\cite{VidMuse} & - & 2024/06 & Diverse Videos & Audio & 360K & 18000 & 180 & Genre \\
        DISCO-MV~\cite{VMAS} & - & 2024/09 & Music Video (DISCO-10M~\cite{DISCO-10M}) & Audio & 2200K & 47000 & 77 & Genre \\
        FilmScoreDB~\cite{HPM} & - & 2024/11 & Film Video & Audio & 32K & 90.3 & 10 & Movie Title \\
        DVMSet~\cite{VidMusician} & - & 2024/12 & Diverse Videos & Audio & 3.8K & - & - & - \\
        HarmonySet~\cite{harmonyset} & \href{https://huggingface.co/datasets/Zzitang/HarmonySet}{Link} & 2025/03 & Diverse Videos & Audio & 48K & 458.8 & 32 & Description \\
        MusicPro-7k~\cite{filmcomposer} & \href{https://huggingface.co/datasets/apple-jun/MusicPro-7k}{Link} & 2025/03 & Film Video & Audio & 7K & - & - & Description, Melody, Rhythm Spots \\

        \midrule
        \noalign{\vskip -1mm}
        \rowcolor{gray!10} \multicolumn{9}{l}{$\blacktriangledown$ \emph{Human Movement Videos}}   \\
        URMP~\cite{URMP} & \href{https://datadryad.org/stash/dataset/doi:10.5061/dryad.ng3r749}{Link} & 2016/12 & Performance Video & MIDI, Audio & 44 & 1.3 & 106 & Instruments \\
        MUSIC~\cite{MUSIC} & \href{https://github.com/roudimit/MUSIC_dataset}{Link} & 2018/04 & Performance Video & Audio & 685 & 45.7 & 239 & Instruments \\
        AIST++~\cite{AIST++} & \href{https://google.github.io/aistplusplus_dataset/download.html}{Link} & 2021/01 & Dance Video (AIST~\cite{AIST}) & Audio & 1.4K & 5.2 & 13 & 3D Motion \\
        TikTok Dance-Music~\cite{D2M-GAN} & \href{https://github.com/L-YeZhu/D2M-GAN}{Link} & 2022/04 & Dance Video & Audio & 445 & 1.5 & 12 & - \\
        \multirow{2}{*}{LORIS~\cite{LORIS}} & \multirow{2}{*}{\href{https://huggingface.co/datasets/OpenGVLab/LORIS}{Link}} & \multirow{2}{*}{2023/05} & Dance Video, Sports Video & \multirow{2}{*}{Audio} & \multirow{2}{*}{16K} & \multirow{2}{*}{86.43} & \multirow{2}{*}{19} & \multirow{2}{*}{2D Pose} \\
        & & & (AIST~\cite{AIST}, FisV~\cite{FisV}, FS1000~\cite{FS1000}) & & & & & \\
        
        \midrule
        \noalign{\vskip -1mm}
        \rowcolor{gray!10} \multicolumn{9}{l}{$\blacktriangledown$ \emph{Images}}   \\
        Music-Image~\cite{Music-Image} & \href{https://mmlab.siat.ac.cn/musicimage_matching/index}{Link} & 2016/07 & Image (Music Video) & Audio & 22.6K & 377 & 60 & Lyrics \\
        Shuttersong~\cite{Shuttersong} & \href{https://dtaoo.github.io/dataset.html}{Link} & 2017/08 & Image (Shuttersong App) & Audio & 586 & - & - & Lyrics \\
        IMAC~\cite{IMAC} & \href{https://gaurav22verma.github.io/IMAC_Dataset.html}{Link} & 2019/04 & Image (FI~\cite{FI}) & Audio & 3.8K & 63.3 & 60 & Emotion \\
        MUImage~\cite{MuMu-LLaMA} & \href{https://huggingface.co/datasets/M2UGen/MUImage}{Link} & 2023/11 & Image (Balanced-AudioSet~\cite{AudioSet}) & Audio & 14.5k & 40.3 & 10 & Instructions \\
        EIMG~\cite{EIMG} & \href{https://github.com/zBaymax/EIMG}{Link} & 2023/12 & Image (IAPS~\cite{IAPS}, NAPS~\cite{NAPS}) & MIDI & 3K & 12.5 & 15 & VA Value \\
        MeLBench~\cite{MeLFusion} & \href{https://schowdhury671.github.io/melfusion_cvpr2024/}{Link} & 2024/06 & Image (Diverse Videos) & Audio & 11.2K & 31.2 & 10 & Genre, Caption \\
        \bottomrule
    \end{tabular}
}
\end{table*}

In this section, we introduce common datasets for vision-to-music. Plenty of datasets have been proposed in the literature of the vision-to-music field, and different methods often use different datasets for training and testing. Therefore, it is necessary to organize and analyze these datasets. Common datasets for vision-to-music generation are listed in Tab.~\ref{tab:datasets}.

\subsection{Input Categories}
Based on the types of videos/images in vision-music datasets, we categorize the datasets as follows:

\noindent\textbf{General Videos.}  
Videos in these datasets are usually sourced from platforms like YouTube. Most datasets focus on Music Videos~\cite{HIMV-200K, MVED, V-MusProd, Video2music, Diff-BGM, VMAS}, as they have satisfactory video-music alignment and are easier to collect, \eg by searching for official music videos using song titles.
However, the diversity of content and styles in these videos may be limited. Other datasets include a variety of video types, such as trailers, advertisements, animations, and documentaries~\cite{VidMusician, VidMuse}, or subsets from larger datasets like AudioSet~\cite{AudioSet}. These videos offer better diversity, but the video-music alignment may be weaker, requiring strict filtering~\cite{MusicCaps, MuMu-LLaMA}. FilmScoreDB and MusicPro-7k~\cite{HPM, filmcomposer} focus on film scores, where the music has a deeper semantic correspondence with the video and serves as an accompaniment rather than being the primary focus, as in music videos.
This provides a unique perspective for the video-to-music generation task, as discussed in Sec.~\ref{sec:fundamentals}. 
Recently, some datasets also provide textual descriptions of videos and music~\cite{MMTrail, MTM, harmonyset} to assist text-bridged video-to-music generation methods.
3.
\noindent\textbf{Human Movement Videos.} 
These videos can be divided into instrument performances and dance/sport categories. 
Representative examples include URMP~\cite{URMP} and MUSIC~\cite{MUSIC}, which aim to reconstruct music from instrumental performance videos. AIST++~\cite{AIST++}, Tiktok Dance-Music~\cite{D2M-GAN}, and LORIS~\cite{LORIS} focus on generating music from dance or sport videos, emphasizing local rhythmic alignment while downplaying semantic relationships.

\noindent\textbf{Images.}  
Existing image-to-music datasets are relatively scarce. Sources of the images are usually frames from music videos~\cite{Music-Image, MeLFusion} or existing image datasets~\cite{IMAC, EIMG, MuMu-LLaMA}. 
Music-Image~\cite{Music-Image} and MeLFusion~\cite{MeLFusion} use frames extracted from online music videos as images. IMAC~\cite{IMAC} and EIMG~\cite{EIMG} select existing image datasets and collect music data, matching images and music based on emotion classification or valence-arousal space. Thus, the image-music pairs are not necessarily one-to-one correspondences. Shuttersong~\cite{Shuttersong} collects user-selected image-music pairs from the Shuttersong application. MUImage~\cite{MuMu-LLaMA} uses images and music from AudioSet, annotated with captions generated by captioning models to provide instructions.

\begin{table}[t]
    \centering
    \caption{Objective metrics of vision-to-music generation. M: MIDI. A: Audio. V: Video. I: Image. T:Text. Pit: Pit. Rhy: Rhythm. Fid: Fidelity. Sem: Semantic.} 
    \vspace{1mm}
    \begin{subtable}{0.9\linewidth}
        \renewcommand\arraystretch{1.1}
        \centering
        \resizebox{\linewidth}{!}{
        \begin{tabular}{llll}
            \toprule
            Metric & Used in Paper & Input & Type \\
            
            \midrule
            \noalign{\vskip -1mm}
            \rowcolor{gray!10} \multicolumn{4}{l}{$\blacktriangledown$ \emph{Music-only:}}   \\
            Scale Consistency & \cite{V-MusProd, S2L2-V2M} & M & Pit \\
            Pit Entropy & \cite{V-MusProd, S2L2-V2M, EIMG} & M & Pit \\
            Pit Class Histogram Entropy & \cite{V-MusProd, XMusic, S2L2-V2M, CMT, Diff-BGM} & M & Pit \\
            Empty Beat Rate & \cite{V-MusProd,XMusic,S2L2-V2M} & M & Rhy \\
            Average Inter-Onset Interval & \cite{V-MusProd, S2L2-V2M} & M & Rhy \\
            Grooving Pattern Similarity & \cite{XMusic,CMT,Diff-BGM} & M & Rhy \\
            Structure Indicator & \cite{CMT, Diff-BGM} & M & Rhy \\
    
            \multirow{2}{*}{Frechet Audio Distance (FAD)} & \cite{VidMuse,MuMu-LLaMA,MeLFusion,V2Meow,MuVi, GVMGen, filmcomposer} & \multirow{2}{*}{A} & \multirow{2}{*}{Fid} \\
            & \cite{MozartTouch,S2L2-V2M,HPM,VEH, audiox, VMAS} & &\vspace{0.6mm} \\
    
            Frechet Distance (FD) & \cite{MTM, VidMusician, SONIQUE, MuVi, VidMuse, MeLFusion, V2Meow, audiox} & A & Fid\vspace{0.6mm} \\
            
            \multirow{2}{*}{Kullback-Leibler Divergence (KL)}  & \cite{VMAS, VidMuse, MozartTouch, MuMu-LLaMA, GVMGen, MTM, VidMusician, V2Meow, MuVi, filmcomposer, HPM} & \multirow{2}{*}{A} & \multirow{2}{*}{Fid} \\
            & \cite{SONIQUE,  MeLFusion, 
 VEH, audiox, S2L2-V2M} & & \\
            
            Beats Coverage Score (BCS) & ~\cite{HPM, MuVi} & A & Rhy \\
            Beats Hit Score (BHS) & ~\cite{HPM, MuVi} & A & Rhy \\
            Inception Score (IS) & ~\cite{HPM, MuVi, audiox} & A & Fid \\
    
            \midrule
            \noalign{\vskip -1mm}
            \rowcolor{gray!10} \multicolumn{4}{l}{$\blacktriangledown$ \emph{Vision-music Correspondence:}}   \\
            ImageBind Score/Rank & \cite{MTM, VidMusician, S2L2-V2M, VidMuse, MozartTouch, MuMu-LLaMA, filmcomposer, audiox} & A,V/I & Sem \\
            CLAP Score & \cite{VidMusician, SONIQUE, VEH} & A,A/T & Sem \\
            Video-Music CLIP Precision & \cite{V-MusProd,S2L2-V2M} & A,V & Sem \\
            Video-Music Correspondence & \cite{Diff-BGM} & A,V & Sem \\
            Cross-modal Relevance & \cite{GVMGen} & A,V & Sem \\
            Temporal Alignment & \cite{GVMGen} & A,V & Rhy \\
            Rhythm Alignment & \cite{VidMusician} & A,V & Rhy \\
    
            \bottomrule
    
            \bottomrule
        \end{tabular}
    }
    \end{subtable}
\label{tab:objective_metric}
\end{table}

\subsection{Music Domains}
Vision-music datasets can be divided into MIDI and audio based on the music modality. MIDI datasets~\cite{V-MusProd, Diff-BGM, Video2music} are created by transcribing audio into the MIDI format or sourced from existing music-only datasets~\cite{Pop909}. Audio datasets contain only raw audio files. 

Compared to audio datasets, MIDI datasets have the following advantages: (1) Include more annotations, such as Chord, Melody, Beat, Tonality, etc; (2) The average duration of each song is longer, enabling the generation of longer music pieces; (3) Suitable for training both symbolic music generation and audio music generation models.
However, a significant limitation of MIDI datasets is their smaller scale (\eg 1K songs, 100 hours vs. 100K-2M songs, 5K-50K hours) and relatively limited diversity. 
For more discussion on the comparison between audio and MIDI (symbolic) data, please refer to Sec.~\ref{sec:challenges}.

\newpage

\section{Evaluation}
\label{sec:evaluation}

Common metrics for vision-to-music are categorized in Tab.~\ref{tab:objective_metric} and ~\ref{tab:subjective_metric}. 
The evaluation of vision-to-music generation can be divided into two categories: \textbf{objective} and \textbf{subjective}. Objective evaluation uses fixed rule-based algorithms or existing models to extract features and calculate musical metrics. 
It is relatively objective and convenient for fair comparison, but has certain biases and cannot cover all aspects of music generation, often differing significantly from human subjective perception. Similar to other generation tasks~\cite{christodoulou2024finding,petsiuk2022human, itinera, rombach2022high}, subjective evaluation is typically used in vision-to-music generation for a more comprehensive assessment, \ie conducting user studies where participants rate/compare music generated by different models. 

From another perspective, metrics can be divided into \textbf{music-only} and \textbf{vision-music correspondence} based on assessment purposes. The former only evaluates whether the music itself is pleasant/realistic/structurally complete, etc., while the latter focuses on the correspondence between the music and the visual input. 

For \textbf{music-only objective metrics}, symbolic music generation methods~\cite{CMT, V-MusProd, XMusic, S2L2-V2M, Diff-BGM, EIMG} use some statistics-based methods to calculate certain pitch or rhythm-related statistical metrics of MIDI, 
such as Scale Consistency, Pitch Entropy, Pitch Class Histogram Entropy, Empty Beat Rate, Average Inter-Onset Interval, Grooving Pattern Similarity, and Structure Indicators. 
These metrics are usually compared with ground truth music, and the closer they are, the more realistic the music is considered. Audio music generation methods widely adopt metrics such as Frechet Audio Distance (FAD), Frechet Distance (FD), and Kullback Leibler Divergence (KL) to 
calculate the similarity between generated music and ground truth music or the distance of predicted class probabilities. 
Some methods~\cite{MuVi, HPM} also introduce metrics like BCS and BHS to measure rhythmic similarity based on music beats.

\textbf{Objective metrics for vision-music correspondence} usually focus on the audio modality. The most commonly used are ImageBind Score/Rank and CLAP score, which leverage pretrained multimodal models like ImageBind~\cite{Imagebind} and CLAP~\cite{CLAP} for similarity evaluation. Some methods~\cite{V-MusProd, S2L2-V2M, Diff-BGM, GVMGen} have also designed specific vision-music retrieval evaluation metrics, with slight differences in model selection and retrieval methods. 
Additionally, GVMGen~\cite{GVMGen} and VidMusician~\cite{VidMusician} have designed objective metrics to evaluate the rhythmic correspondence between visions and music. 
However, since the pretrained models are usually trained with general audio data instead of specified music data, these objective metrics commonly do not perfectly align with human judgments.

\textbf{Subjective metrics} mainly include MOS (generally using a 5-point Likert scale), pair preference (\ie win rate), and ranking different music. Common subjective metrics in vision-to-music generation are given in Tab.~\ref{tab:subjective_metric}.
Among these, music-only metrics mainly include overall music quality, music richness, music melody, and music rhythm, while vision-music metrics include overall correspondence, semantic consistency, rhythm consistency, and emotion consistency. 
The selection of specific subjective metrics depends on the vision-music relationship emphasized by the method.

\begin{table}[t]
    \caption{Subjective metrics of vision-to-music generation.}
    \vspace{1mm}
    \centering
    \begin{subtable}{\linewidth}
        \renewcommand\arraystretch{1.3}
        \centering
        \resizebox{0.8\linewidth}{!}{
        \begin{tabular}{ll}
            \toprule
            Metric & Used in Paper \\
            
            \midrule
            \noalign{\vskip -1mm}
            \rowcolor{gray!10} \multicolumn{2}{l}{$\blacktriangledown$ \emph{Music-only:}}   \\
            Music Melody & \cite{V-MusProd,Diff-BGM} \\
            Music Rhythm & \cite{V-MusProd,Diff-BGM} \\
            Music Richness & \cite{GVMGen, XMusic} \\
            Audio Quality & \cite{VidMuse, MuVi} \\
            Overall Music Quality & \cite{GVMGen,M2M-Gen,VMAS,Video2music,MozartTouch,MeLFusion,V2Meow,VEH,MTM,VidMuse,CMT,filmcomposer} \\
    
            \midrule
            \noalign{\vskip -1mm}
            \rowcolor{gray!10} \multicolumn{2}{l}{$\blacktriangledown$ \emph{Vision-music Correspondence:}}   \\
            Semantic Consistency & \cite{V-MusProd,MTM,VidMusician, MuVi, Diff-BGM,filmcomposer} \\
            Rhythm Consistency & \cite{V-MusProd,XMusic,MTM,VEH,Diff-BGM,Video2music,filmcomposer} \\
            Emotion Consistency & \cite{XMusic,MTM,VEH} \\
            Overall Correspondence & \cite{V-MusProd,SONIQUE,M2M-Gen,VMAS,S2L2-V2M,VidMuse,CMT,Video2music,MozartTouch,MuMu-LLaMA,MeLFusion,V2Meow,GVMGen,filmcomposer} \\
            \bottomrule
        \end{tabular}
    }
    \end{subtable}
\label{tab:subjective_metric}
\end{table}

\vspace{-1mm}
\section{Discussion and Challenges}
\label{sec:challenges}

\subsection{Popular Research Topics}

Early work in vision-to-music generation focused on rule-based relationships and symbolic music paradigms. In the past couple of years, however, rapid advances in multimodal large models, audio generation, and related data developments have spurred significant progress in the field.
Several emerging research directions aiming to improve quality, alignment, and real-world usability. Specifically, current popular research topics include:

\begin{enumerate}[label=\arabic*.,leftmargin=*]
\item \textbf{Multimodal Learning Architectures:}
Researchers are designing new architectures that better connect visual and musical modalities. This includes the use of transformers, cross-attention mechanisms, and shared embedding spaces to capture complex visual cues and translate them into music features.
\item \textbf{Emotion and Rhythm Alignment:}
Mapping visual emotion (e.g., joy, tension) and movement patterns to corresponding music is a key challenge. Ongoing work explores how to extract affective and temporal signals from videos and align them with musical attributes such as tempo, tonality, and dynamics.
\item \textbf{Dataset Development and Evaluation:}
Creating high-quality, diverse vision-music datasets remains a bottleneck. Researchers are also working on better evaluation metrics that align with human judgment, as existing ones often fail to capture subjective musical quality and visual-musical correspondence.
\end{enumerate}

These topics reflect a broader shift from purely technical generation to more expressive, controllable, and user-aligned music synthesis, opening doors to creative and industrial applications.

\subsection{Challenges}

Despite these advances, we identify several key challenges for the academic community:

\noindent\textbf{Lack of Standardized Objective Datasets and Benchmarks:} 
The training and evaluation datasets differ across models, sometimes leading to comparisons between models fine-tuned on proprietary datasets and those evaluated via zero-shot inference on other datasets. This disparity significantly undermines the fairness of model comparisons and makes it challenging to identify the true state-of-the-art (SoTA). Besides, current evaluation metrics often do not align with actual human perception, \eg FAD and KL are trained on general audio data rather than on music-specific data, and symbolic metrics are statistically based and exhibit low correlation with human preferences. Though most papers provide demos for qualitative comparisons, they are prone to issues such as cherry-picked examples, insufficient sample size, and subjectivity in evaluating the outputs.

\noindent\textbf{Limited Customization and Controllability:}
Most existing models function as black boxes, making it challenging to personalize or control attributes of the generated music, such as style, instrumentation, and rhythm. This significantly affects the models' practical applicability. 
Future research needs to focus on aligning generated music more closely with human preferences.
    
\noindent\textbf{Trade-off Between Symbolic and Audio Forms:}
As discussed in Sec.~\ref{sec:fundamentals}, audio-based methods benefit from large-scale data but generally offer limited controllability and are constrained by the computational cost of high-fidelity generation, often resulting in shorter musical pieces. In contrast, symbolic approaches, while limited by available data, offer better controllability and can produce longer compositions. A promising direction is to combine symbolic and audio methods to achieve a better trade-off. 
Integrating certain musical priors can enhance data efficiency; however, relying solely on rule-based approaches may limit the model's capabilities, which is a trade-off that merits deeper exploration.

\noindent\textbf{Cross-Domain Generalization:}
While current vision-to-music systems often perform well within specific datasets or contexts, their ability to generalize across different domains (e.g., film, video games, YouTube videos) remains limited. Research is focusing on developing robust models that can generalize across a wide range of visual and musical genres, ensuring that the generated music is coherent, contextually appropriate, and stylistically diverse. 
This involves training models on diverse, large-scale datasets and developing transfer learning techniques that allow models to adapt to new contexts without extensive retraining. Ensuring generalization will be essential for the scalability and real-world applicability of vision-to-music systems, particularly as they move toward industrial applications.

\clearpage

\noindent\textbf{Limited Human-in-the-Loop Feedback for Iterative Improvement:}
Currently, most vision-to-music systems rely solely on offline datasets and predefined evaluation metrics, lacking active human involvement in the model development cycle. However, music is an inherently creative and subjective domain, where subtle aspects such as emotional progression, stylistic nuance, and narrative tension are difficult to capture with automated systems alone. Some of them are even challenging to distinguish for most people and require professionals to participate. Therefore, incorporating \emph{human-in-the-loop} frameworks — especially involving professional composers, musicians, or even target audiences — can offer several advantages. First, expert composers can provide iterative feedback during model training, guiding the model to learn more expressive and contextually appropriate musical patterns. Second, interactive fine-tuning systems can allow users to steer generated results through simple feedback mechanisms (e.g., tags and critiques), enabling personalization and controllability. Third, human-curated data and active learning strategies can prioritize ambiguous or low-confidence samples for labeling or correction, improving both data quality and model alignment. Encouraging deeper collaboration between machine learning researchers and music professionals will be key to building systems that are not only technically competent, but also artistically compelling and socially acceptable.

\textbf{Under-utilization of Large Model Capabilities:}
The vision-to-music field, or even the broad music generation area has yet to experience a breakthrough akin to the ``GPT moment.'' Despite inspiring commercial produces and open-source research like Suno\footnote{https://suno.com} and YuE~\cite{yuan2025yue}, he current growth in data and model capabilities has not reached a qualitative turning point. Achieving a broadly applicable foundational model for music generation will require substantial investments in resources and data, particularly for foundational tasks such as unconditional music generation or text-to-music synthesis.

Beyond these academic challenges, a compelling area of exploration is to aligning these technologies with the demands of the music industry and end users --- whether for consumer products (such as automatically generated background music on short video platforms) or professional applications (such as film and game scoring) --- presents significant commercial opportunities.

\subsection{Applications}

Vision-to-music generation holds significant promise across a wide range of real-world applications. In the entertainment industry, it can streamline film scoring, game music generation, and TV production, enabling composers to rapidly prototype background music tailored to visual narratives, thereby improving creative workflows and reducing production cycles. In short video platforms (e.g., TikTok, Instagram Reels), automatic music generation based on visual input allows content creators to avoid copyright issues and personalize audio tracks, leading to a richer and more engaging user experience. For virtual reality (VR) and augmented reality (AR) environments, dynamically generated background music synchronized with the user’s visual experience can enhance immersion and emotional impact. In the realm of personal media, such as photo albums, home videos, or social media posts, vision-to-music systems can automatically generate soundtracks, making content more compelling and expressive. Moreover, in education and therapy, the technology can assist in music therapy, creative learning, or even help individuals with visual impairments experience visuals through music, thus extending accessibility. As these applications mature, aligning technical capabilities with user needs will be critical for successful integration into real-world products.

\vspace{-1mm}
\section{Conclusion}

Vision-to-music generation represents a promising yet relatively underexplored frontier in the multimodal artificial intelligence landscape. By bridging visual content with musical creation, this task holds potential across creative industries, from film and gaming to short video platforms and personal media. In this survey, we provided a comprehensive overview of the field, covering input/output modalities, technical approaches, architectural designs, vision-music relationships, datasets, and evaluation methods. 
We highlighted key challenges such as aligning visual cues with expressive music, balancing symbolic and audio generation, and the absence of standardized benchmarks. Looking ahead, progress will hinge on large multimodal models, better datasets, and human-in-the-loop feedback, enabling more controllable and generalizable systems for real-world applications.
We hope this survey serves as a foundation for further exploration and can pave the way for future research and applications on vision-to-music generation.

\bibliographystyle{plain}
\bibliography{main}

\end{document}